\documentclass[letterpaper, 10 pt, conference]{template/ieeeconf}  

\IEEEoverridecommandlockouts                              
\usepackage{graphics} 
\usepackage{epsfig} 
\usepackage{mathptmx,cuted} 
\usepackage{times} 
\usepackage{amsmath} 
\usepackage{amssymb}  
\usepackage{hyphenat}
\usepackage{textcomp}
\usepackage[tight,footnotesize]{subfigure}
\long\def\invis#1{}

\def\floatsep{4.0pt}
\def\textfloatsep{0.0pt}
\def\dblfloatsep{4.0pt}
\def\dbltextfloatsep{4.0pt}
\title{\LARGE \bf Dynamically Efficient Kinematics for Hyper-Redundant Manipulators}

\author{Marios Xanthidis$^{1}$, Kostantinos J. Kyriakopoulos$^{2}$, and Ioannis Rekleitis$^{1}$%
\thanks{$^{1}$Marios Xanthidis and Ioannis Rekleitis are with the Computer Science and Engineering Department, University of South Carolina,
United States, {\tt\small mariosx@email.sc.edu,yiannisr@cse.sc.edu}}%
\thanks{$^{2}$Kostantinos Kyriakopoulos is with the Department of Mechanical Engineering, National Technical University of Athens, Greece, {\tt\small kkyria@mail.ntua.gr}}%
}
\begin{document}

\maketitle
\thispagestyle{empty}
\pagestyle{empty}

\begin{abstract}
A hyper\hyp redundant robotic arm is a manipulator with many degrees of freedom, capable of executing tasks in cluttered environments where robotic arms with fewer degrees of freedom are unable to operate. This paper introduces a new method for modeling those manipulators in a completely dynamic way. The proposed method enables online changes of the kinematic structure with the use of a special function; termed ``meta\hyp controlling function''. This function can be used to develop policies to reduce drastically the computational cost for a single task, and to robustly control  the robotic arm, even in the event of  partial damage. The direct and inverse kinematics are solved for a generic three-dimensional articulated hyper\hyp redundant arm, that can be used as a proof of concept for more specific structures. To demonstrate the robustness of our method, experimental simulation results, for a basic ``meta\hyp controlling'' function, are presented.
\end{abstract}

\section{Introduction} 
\label{sec_intro}

Robotic arms are one of the most utilized products of robotics, with both scientific and industrial applications thanks to their efficiency, their accuracy, and their ability to work non-stop in challenging environments. They consist of sequential links that have the ability to move with respect to the previous link, in a way that the structure could be described by a kinematic chain. In literature, a kinematic chain is often categorized according to the Degrees of Freedom (DoFs). Arms with 3 DoFs are able to move in the 3-dimensional space, while arms with 6 DoFs are able to both move and rotate in the 3-dimensional space. Arms with 7 DoFs mimic the human arm and when the degrees of freedom are more than 6, they are termed hyper\hyp redundant. Hyper\hyp redundant arms use the extra DoFs for maneuvers in  space when there are obstacles that block the motion for ordinary arms. As such, arms with many or even infinite degrees of freedom, \emph{hyper\hyp redundant robotic arms}, are ideal for operating in complex environments where the free space is limited~\cite{a55}; such as cluttered environments.

\invis{The algorithm that controls a robotic arm is called manipulator and it is responsible for moving the end effector to the desired position and orientation by moving every link of the arm in the proper position. }
Robotic arms (manipulators) have one end fixed and the other end, usually carrying an instrument or a gripper, moving through free space; the free moving end is called the end effector. Inverse kinematics algorithms are responsible for moving the end effector from the start pose (position and orientation) to the goal pose. Moving the end effector to the goal pose requires to move each link of the manipulator through free space without any collisions. Finding the required motions in an efficient manner can be cast as a minimization problem and there are optimal solutions for non\hyp redundant arms and near optimal for redundant arms~\cite{a22}. \invis{To solve this problem it is necessary to solve two separate problems, the \emph{direct} and the \emph{inverse kinematic} problems. The direct kinematic is the problem of finding the end-effector's pose if the position of the other links is known. The inverse kinematics is the problem of finding the optimal positions for every joint so the end-effector reaches a desired pose. For every task every manipulator has to find the current position of the end-effector by solving the direct kinematics problem, finding the difference to the current and the desired position, then solve the inverse kinematics problem, and finally move the end effector, iteratively until the end-effector reach the desired goal.}

Direct kinematics is the analytical solution that calculates the end\hyp effector pose for a given arm configuration; inverse kinematics is the calculation that defines the configuration of the arm to achieve a desired end\hyp effector pose. The standard formulation for the direct kinematics can be explained by considering the robotic arm as a sequence of links and joints starting from its base and ending up to its end-effector. Every link has the ability to move with respect to the previous link, and such a kinematic structure is also known as a \emph{kinematic chain}. Each link's pose is described by a transformation matrix with respect to the previous link. Given the transformation matrices of all the links of the manipulator, the pose of the end-effector, namely the direct kinematics, can be easily calculated by multiplying all transformation matrices, starting from the first link until the last one. The analytical solution is the same for both robotic arms and hyper\hyp redundant manipulators.

The formulation of the inverse kinematics problem is done by the so called \emph{differential kinematics}, when the robotic arm has a complex kinematic structure. In such a formulation, the relation between the motion of each link and of the end-effector is described by a Jacobian matrix. Then, the inverse of the Jacobian matrix is multiplied with the desired linear and angular velocities of the end-effector to get the necessary velocities of each joint. When the dynamic constraints of the manipulator are not considered, the velocity space trivially maps to the pose space. In most cases the Jacobian matrix of redundant structures is not invertible, so a pseudo-inverse matrix is used instead, produced by the damped least squares method \cite{nakamura1986inverse,wampler1986manipulator}. 

This paper proposes an analytical solution on the direct and the inverse kinematics problems for hyper\hyp redundant manipulators, by modeling the problems in a dynamical way. The proposed method provides the ability to not only control the robotic arm but also its kinematic structure, to reduce the complexity by reducing the unnecessary DoFs. In addition, structure failures due to damaged links, are also handled by the discussed algorithm. The remainder of this paper is structured as follows,  related work and the traditional approach to these problems is discussed in section \ref{sec:bg}. The proposed solution with a simple application are presented in section \ref{sec:solution}. Section \ref{sec:results} demonstrates the efficiency of the proposed method with simulated experimental results. Finally, the paper concludes with lessons learned and a discussion of future work in section \ref{sec:concl}.

\section{Background}
\label{sec:bg}
\subsection{Related Work}
Hyper\hyp redundant robotic arms are useful, for operations in extremely cluttered environments with only a small free operational space as demonstrated by Ma et al.~\cite{a10} who developed a hyper\hyp redundant arm to make real time, precise operations inside nuclear reactors. Liljeback et al.~\cite{a2} created a snake fire-fighting  hyper\hyp redundant robot, and Ikuta et al.~\cite{a6} used a 9-DoF arm to operate during surgeries in deep areas.

There are many technical difficulties arising from having big open-chained kinematic structures because of the exponential increment of the splines forces at the joints to all the body. This problem prohibited the application of the first hyper\hyp redundant manipulators in three-dimensional space. For example studies such as the one by Chirikjian and Burdick~\cite{a5} used a 2 dimensional arms with 30 degrees of freedom. There are some solutions that have been proposed addressing some of those technical difficulties such as studies that focus on building hyper\hyp redundant robotic arms~\cite{a17,a18,a7,a4,a3,a16}, and studies that deal with the design of joints with many DoFs per joint, used for hyper\hyp redundant arms \cite{a19,a13,a14}.
\invis{by Qiang et al.~\cite{a17}, Hirose et al.~\cite{a18}, Brown et al.~\cite{a7}, Sujan et al.~\cite{a4}, and Ning et al.~\cite{a3}, and Chibani et al.~\cite{a16} that focus on building hyper\hyp redundant robotic arms. Other studies by Behrens et al.~\cite{a19},  Shammas et al.~\cite{a13} and~\cite{a14} deal with the structure of joints of many DoFs that used for hyper\hyp redundant arms.}

Leaving aside the technical limitations there are also problems deriving from the complexity of the kinematic and dynamic models of those arms. The complexity issues have been extensively discussed by Chirikjian~\cite{a55} where some macroscopic curve fitting solutions have been imposed. The complexity of dealing with hyper\hyp redundant arms, according to previous studies, is a problem that is treated as four subproblems: The dynamics that aim to control the robotic arm more efficiently with respect to the forces the robotic arm accepts and exerts. Improvements in the kinematics which aim  to compute more efficiently the kinematic chain and control the actuators. Obstacle avoidance methods which aim to move the hyper\hyp redundant manipulator in environments with obstacles in an optimal manner. Fourth fault tolerance mechanisms which aim to control a hyper\hyp redundant robotic arm even when it is damaged.

For the dynamics problem a first continuum approximation approach with the backbone curve and the ability to parallelize the problem was introduced by Chirikjian~\cite{a53}. Other ways to solve the dynamics problem for a hyper\hyp redundant arm have been presented by Ma et al.~\cite{a49}. In particular, a dynamic control scheme is introduced in a constrained operational space. In addition with a dynamic formulation for the hyper\hyp redundant robotic arm, the solution is provided efficiently and the execution is accurate. Moreover, other approaches as presented in the work of Oda et al.~\cite{a48} aimed to design manipulators by focusing on the desired force at the end-effector to achieve a desired acceleration. Other studies use biomimetic approaches to solve efficiently the problem (Godage et al.~\cite{a51}, and Kang et al.~\cite{a50}) that consider also the hydrodynamic forces in underwater applications, and also studies based on the screw theory can be seen in the work of Gallardo et al.~\cite{a52}.

For the kinematics, which is the problem we are focusing, the pioneering work by Chirikjian and Burdick~\cite{a36} focused on modeling and controlling efficiently hyper\hyp redundant arms and providing obstacle avoidance algorithms. They showed that seeing the macroscopic geometrical features of those arms and using the backbone curve technique, leads to describing the structure with less variables and reduces the computational needs. With this technique the effective parallel computation for each backbone curve is introduced; then some numerical improvements in the inverse kinematics~\cite{a31} and formulation followed~\cite{a34} that led to near optimal motion planners~\cite{a35}. Other near optimal planners, that also use some points of the structure to determine an optimal shape of the hyper\hyp redundant arm, have been presented from Zanganeh and Angeles~\cite{a29}. Another approach by Kobayashi et al.~\cite{a27} tries to solve the problem by defining first the shape for the robotic arm and then  each joint is evaluated to satisfy the desired shape. Some other techniques for reducing the computational cost, as the one proposed by Ebert-Uphoff and Chirikjian~\cite{a33}, assume that every joint has discrete and finite possible positions. So by using a tree structure they reduced the computations for the results, since there is no need for solving the inverse kinematics in the classical analytical differential way. Some studies by Fahimi et al.~\cite{a32} combine the curve backbone concept with the discrete joints assumption with some new mode shapes, to offer a velocity solution for spatial hyper\hyp redundant arms. Other geometrical approaches by Yahya et al.~\cite{a25,a28} offer near optimal solutions, and show that setting the angles of the adjacent joints equal leads to computationally efficient results, and provides more stable behavior of the arm. Another solution by Nearchou~\cite{a26} has a higher level approach and uses genetic algorithms to find fast solutions without the need of computing the pseudo-inverse for the inverse kinematics. Last but not least, dynamic models for the inverse kinematics have been presented by Wang et al.~\cite{a30} who show that the solutions do not depend on the number of joints but depend on some virtual segments. This finding allows to reduce the variables of the problem and thus the computational needs.

Several algorithms in addition to the ones discussed above have been proposed for solving the obstacle avoidance problem \cite{a35,a29,a32,a28,a26}. Other techniques use the idea of considering the free space as tunnels in the obstacle filled workspace and use differential kinematics as done by Chirikjian and Burdick~\cite{a41,a42}; while other techniques, as presented by Ma and Konno~\cite{a40}, separate the desired path into close points and for each step the manipulator tries to avoid the obstacles. Recent studies by da Gra$\c{c}$a et al.~\cite{a39} combine the closed-loop differential pseudo-inverse way of the methods presented above, with genetic algorithms that provide repeatability for closed path operations.

In contrast to regular robotic arms, a hyper\hyp redundant manipulator can be functional even if it has many damaged joints, so it is essential to be able for a controller to control the manipulator properly with or without damage. An approach that explicitly deals with failures is that of Kimura et al.~\cite{a45}, who presents decentralized autonomous architecture for space hyper\hyp redundant manipulators.

This paper, in contrast with the ones already presented, introduces an algorithm that does not make any macroscopic assumptions, other than the Jacobian matrix, and solves the problem in an analytical way. The proposed method is without the limitation of applying only one method for reducing the DoFs, and the manipulator has the ability to change its kinematic structure in the process. Moreover, not only the computational cost is unrelated to the size of the hyper\hyp redundant arm, but also the ability of developing many policies by combining different geometrical methods is provided.

\subsection{The Classic Method}
Next the classical method is presented, since it is used as a building block of our method. Furthermore we show that our method does not have the limitations of the classic formulation. As discussed in the previous section, in the direct kinematics the aim is to describe the pose of the end-effector as a homogeneous transformation with respect to the base. The homogeneous transformation for every joint is expressed with respect to the previous joint. For the $i^{th}$ joint the homogeneous transformation with respect to the previous $i-1$ joint can be given by the following matrix \cite{a22}:
\begin{equation}
\begin{split}
A^{i-1}_i(q_i)=
\begin{bmatrix}
    n^{i-1}_i(q_i)  & s^{i-1}_i(q_i)   & a^{i-1}_i(q_i)     & p^{i-1}_i(q_i) \\
    0           & 0            & 0              & 0 
\end{bmatrix}\implies \\ 
A^{i-1}_i(q_i)=
\begin{bmatrix}
    R^{i-1}_i(q_i)     & p^{i-1}_i(q_i) \\
    0                 & 0 
\end{bmatrix}
\end{split}
\end{equation}
where $R^{i-1}_i$ is the rotation matrix, $n^{i-1}_i$, $s^{i-1}_i$, $a^{i-1}_i$ are the rows of the $R^{i-1}_i$, $p^{i-1}_i$ the translation matrix, and $q_i$ the variable related to the motion of the $i$ joint. For simplicity's sake, in the following $A^{i-1}_i$ is denoted as $A_i$. Using the multiplication for an open chain manipulator of $n$ joints the transformation matrix for the pose of the end effector $e$ with respect to the base frame $b$ is:
\begin{equation}
T^b_e(q)=
\begin{bmatrix}
    n^b_e(q)  & s^b_e(q)   & a^b_e(q)     & p^b_e(q) \\
    0           & 0            & 0              & 0 
\end{bmatrix}
={\displaystyle \prod_{i=1}^{n} A_i(q_i)}
\end{equation}
where $q=[q_1, q_2,..., q_n]^T$.

In inverse kinematics, to describe how every joint affects the pose of the end-effector the Jacobian matrix is needed. The Jacobian matrix for an articulated robotic arm with revolute joints is give by the following equation:
\begin{eqnarray}
\label{JJJ}
&J=
\begin{bmatrix}
    J_{P_{1}}  & J_{P_{2}}   & \dots     & J_{P_{n-1}}   & J_{P_{n}} \\
    J_{O_{1}}  & J_{O_{2}}   & \dots     & J_{O_{n-1}}   & J_{O_{n}}
\end{bmatrix},\\
&\begin{bmatrix}
    J_{P_{i}} \\
    J_{O_{i}}
\end{bmatrix}
=
\begin{bmatrix}
     a_{i-1}\times(p_e-p_{i-1})\\
    a_{i-1}
\end{bmatrix}
\end{eqnarray}
To solve the inverse kinematics problem the following pseudo-inverse Jacobian can be used:
\begin{equation}
J^*=J^T(JJ^T+k^2I)^{-1}
\end{equation} 
  where $k$ is a damping factor from the implementation of Damped Least Squares method to the following cost function: 
\begin{equation}
g(\dot{q})=\frac{1}{2}(u_e-J\dot{q})^T(u_e-J\dot{q})+\frac{1}{2}k^2\dot{q}^T\dot{q}
\end{equation}
 where $u_e$ is the vector that includes the desired linear and angular velocity of the end-effector to go from the current pose to the desired pose. Finally the solution to the inverse kinematics problem is:
\begin{equation}
\label{qtk}
q(t_k)=q(t_{k-1})+\dot{q}(t_{k-1})*\Delta t
\end{equation} 
where $\Delta t$ is the duration between the $t_k$ and $t_{k-1}$ moments and $\dot{q}(t_k)$ is given by the following function:
$\dot{q}=J^*u_e$

\section{Dynamic method}
\label{sec:solution}
\subsection{Description of the method}
The classic method of modeling efficiently a robotic arm may seem robust for ordinary arms, but it encounters two big problems when it comes to hyper\hyp redundant structures:

First, to control a robotic arm of $n$ DoFs, for every step, the following operations are necessary: $n$ $4\times4$ matrix multiplications, construction of the Jacobian $6\times n$ matrix and the pseudo-inverse, and then solving Equation (\ref{qtk}). The complexity of a manipulator is quadratic on the number of the degrees of freedom of its structure. This is not a real problem for structures with few degrees of freedom (regular robotic arms) but it is unmanageable for structures with many DoFs, as the hyper\hyp redundant manipulators.

Second, the classic method does not provide any tools for controlling a broken robotic arm, since a robotic arm of 6 DoFs, with at least one broken joint, cannot operate in the 3D space, but a hyper\hyp redundant arm is still functional if it has more than or equal to 6 functional joints.

The proposed solution for the first problem is to reduce dynamically the total degrees of freedom of the structure. The idea of the proposed algorithm could be summarized in the following sentence: It is unlikely to need the hyper\hyp redundant arm to be fully operational and to need all of its kinematic abilities in ordinary environments and tasks.
The solution for the second problem can be found by separating the kinematic structure into a group of functional links and a group of links that are damaged, and exclude the second group from the Jacobian matrix see Equation (\ref{JJJ}). The proposed solution in this paper provides a formulation that combines those two ideas. 
A hyper\hyp redundant arm is separated into consecutive sectors, where every sector uses a subset of $q$. In every sector the homogeneous transformation (with respect to the previous sector) is produced by the sequential multiplications of only the homogeneous transformations matrices that are related to these DoFs and are member of that subset. Finally the transformation matrix of the end-effector, with respect to the base is achieved by the consecutive matrix multiplication of the final products of every sector. More precisely, let $q$ be the vector of the $n$ joint variables, $q_t$=$[{q_{m_{t-1}+1}, q_{m_{t-1}+2}, ... q_{m_t}}]$ be the subset of $q$ corresponding to the $t$ sector with $m_t$ links, then for a structure of $r$ sectors:
\begin{equation}T^b_e(q)={\displaystyle \prod_{t=1}^{r}{\displaystyle \prod_{j=1}^{m_t} A_{m_{t-1}+j}(q_{m_{t-1}+j}) }}\label{ppp}\end{equation}
\noindent Note that $m_0=0$.

The computational cost of the multiplication in Equation (\ref{ppp}) can be reduced by using only a subset $Q$ of the available DoFs of $q$ and then applying an efficient method inside every sector $t$, such as a geometric one, to find the homogeneous transformations $A_{t_i}$ between the joints that respond to the $l_t$ variables in the subset $Q_t=[q_{t_{1}}, q_{t_{2}},... ,q_{t_{l_t}}]^T$ of $q_t$. Then, the result of this reduction would be the following double product:
\begin{eqnarray}&T^b_e(Q)={\displaystyle \prod_{t=1}^{r}{\displaystyle \prod_{i=1}^{l_t} A_{t_i}(q_{t_i}) }}\text{    and}\label{Tr}\\
&Q= 
\begin{bmatrix}
     Q_1^T & Q_2^T & \dots & Q_r^T
\end{bmatrix}^T\end{eqnarray}

Similarly the Jacobian matrix can be constructed by computing the product of the homogeneous transformation of the previous sectors and the homogeneous transformation product inside the sector until the transformation matrix of the current joint. So the Jacobian has the following form:
\begin{eqnarray}
&J=
\begin{bmatrix}
     J_1 & J_2 & \dots & J_t & \dots & J_r
\end{bmatrix},\\
&J_t=
 \begin{bmatrix}
     J_{P{t_1}}     & J_{P{t_2}}    & \dots     & J_{P{t_l}} \\
     J_{O{t_1}}     & J_{O{t_2}}    & \dots     & J_{O{t_l}}
\end{bmatrix}
\end{eqnarray}
To achieve the above formulation it is necessary to use some helpful data structures that would store the important information of our structure, all need for calculating the transformations in Equation (\ref{Tr}). They would also provide the relations between the real kinematic structure of the robotic arm and the current kinematic structure based on the sectors modeling. More specifically, we introduce three matrices ($V$, $F$, $H$) which store the necessary informations. The $V$ matrix contains a set of useful information for every sector ($t_1, \ldots, t_l$, $q_{t_1}, \ldots, q_{t_l}$, etc.). The $F$ matrix that contains the damaged links. The H matrix contains the kinematic correspondence for every link of the structure. By simply using the values of the $H$ matrix, the $V$ and the $F$ matrices can be constructed and the direct and inverse kinematics problem can be solved, with the right correspondence between the configuration $q$ and the reduced configuration $Q$.

The H matrix contains an identification number that describes the kinematics for a single link that may contains several joints; each different value corresponds to a different homogeneous transformation for this link. The value for every link is determined by the meta\hyp control function, or meta\hyp controller, $h$. So, for a robotic arm of $n$ links and $f$ different abilities of movement or combinations of those abilities for a link, $h$ is defined as:
\begin{equation} h:\{1,2,3,...n\}\rightarrow\{-1,0,1....f-1\}\end{equation}
\noindent where \{-1\} corresponds to a broken link. By definition the meta\hyp controller can change at any time the current kinematic structure and it is the key function to the new fully dynamical approach introduced in this paper. Every manipulator, to operate in the most difficult environments and tasks, should be able to use all the available DoFs if needed.

In the next subsection an example is presented to show how our method can be applied to a generic hyper\hyp redundant manipulator and how under certain assumptions the complexity of the problem can be decreased significantly.

\subsection{Implementation example}
\label{sec:impl_ex}
\begin{figure}
\centering
\begin{tabular}{cc}
     \subfigure[]{\includegraphics[width=0.27\textwidth]{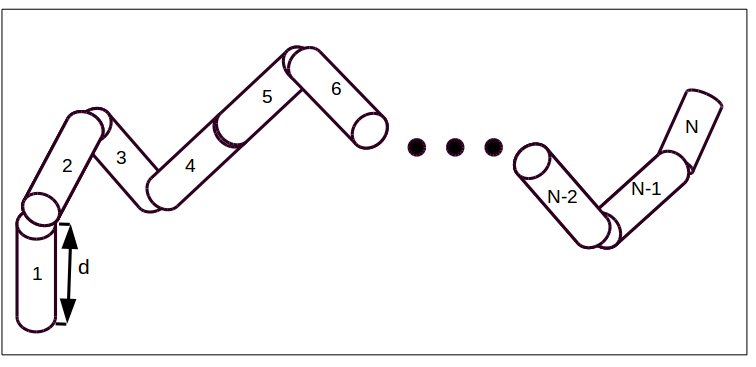}}&
     \subfigure[]{\includegraphics[width=0.19\textwidth]{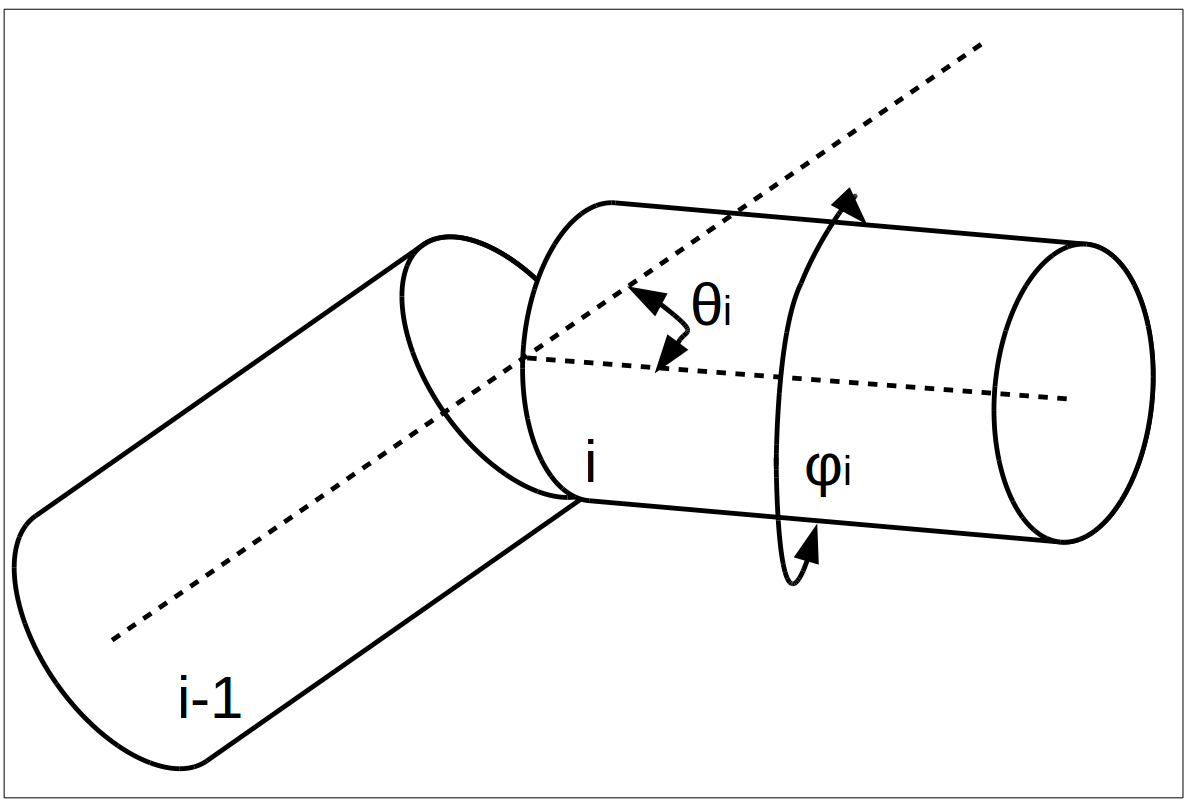}}
\end{tabular}
\caption{(a) The structure of the generic hyper\hyp redundant arm of $N$ links with length $d$. (b) The two DoFs link between two links.} 
\label{fig_arm}
\end{figure}

\invis{
\begin{figure}
\centering
\includegraphics[width=0.25\textwidth]{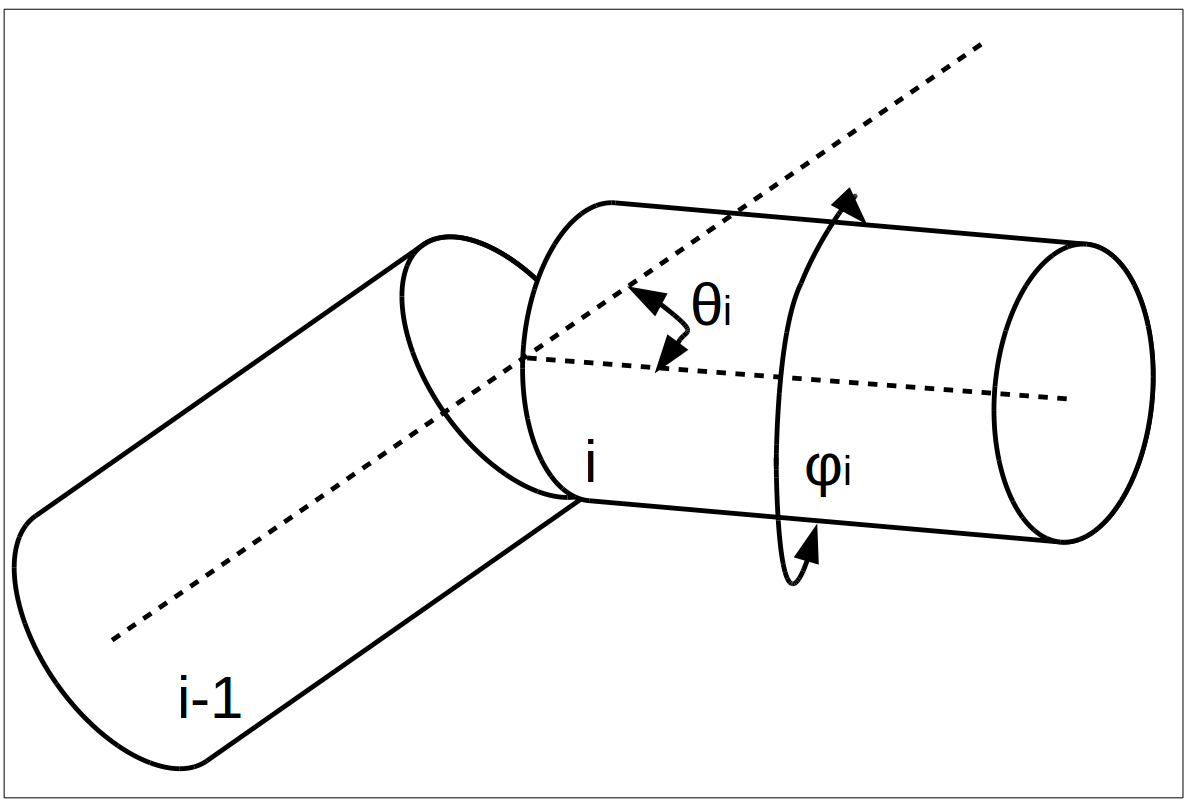}
\caption{The two DoFs link between two links.} 
\label{fig_link}
\end{figure}
}
In this section the process of construction of the $V$, $F$, and $H$ matrices, the solution of both the direct and inverse kinematics, and finally a simple meta\hyp controller.

A generic hyper\hyp redundant arm is composed of $N$ cylindrical links, for simplicity sake, of the same length d, where every link has 2 DoFs; see Fig.\ref{fig_arm}(a). It can bend with respect to the previous link by $\theta$ and rotate with respect to its cylindrical axis by $\phi$; see Fig. \ref{fig_arm}(b). So $q=[\phi_1, \theta_1,  \phi_2, \theta_2, ..., \phi_N, \theta_N]^T$, by applying the classic method, the transformation of the end-effector is:
\begin{eqnarray}
&T^b_e(q)={\displaystyle \prod_{i=1}^{N} A_{1_i}(\phi_i)A_{2_i}(\theta_i)},\\
&A_{1_i}(\phi_i)=
 \begin{bmatrix}
     \cos(\phi_i)       & \sin(\phi_i)    & 0   & 0 \\
     -\sin(\phi_i)     & \cos(\phi_i)    & 0   & 0 \\
     0                  & 0               & 1   & 0 \\
     0                  & 0               & 0   & 1
\end{bmatrix}\\
&A_{2_i}(\theta_i)=
 \begin{bmatrix}
     1                  & 0                 &0                  & 0 \\
     0                  & \cos(\theta_i)    & \sin(\theta_i)    & d\sin(\theta_i) \\
     0                  & -\sin(\theta_i)   & \cos(\theta_i)     & d\cos(\theta_i) \\
     0                  & 0                 & 0                 & 1
\end{bmatrix}
\end{eqnarray}

\noindent and
\begin{eqnarray}
&J=
\begin{bmatrix}
     J_1 & J_2 & \dots & J_i & \dots & J_N
\end{bmatrix},\\
&J_i=
 \begin{bmatrix}
     J_{P_{1_i}}    &J_{P_{2_i}}\\
     J_{O_{1_i}}    &J_{O_{2_i}}
\end{bmatrix}\end{eqnarray}
\begin{figure}
\centering
\includegraphics[width=0.35\textwidth]{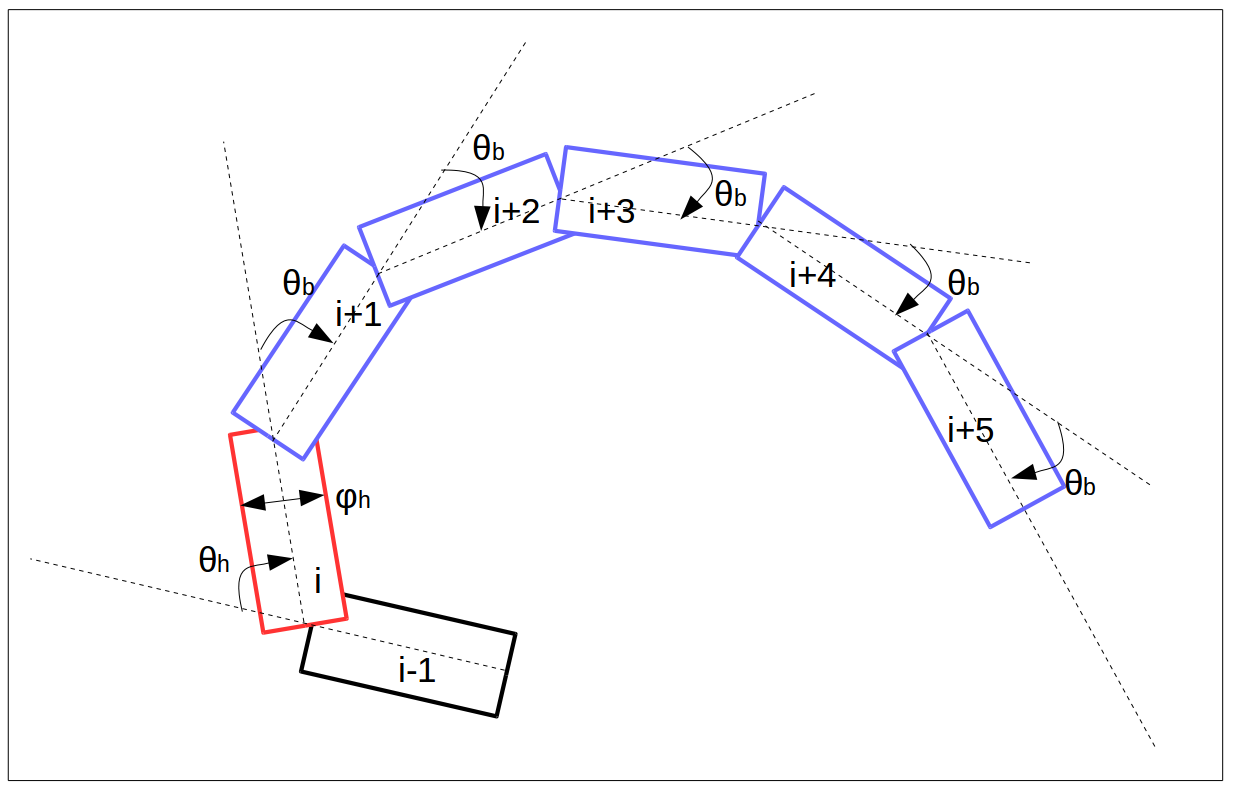}
\caption{A sector of 6 links and its geometrical structure. The head (i) is shown with red color, the links of the body (i+1:i+5) with blue, and with black the other links. } 
\label{fig_sector}
\end{figure}
Back to the proposed method, the definition of the properties of the structure of a sector should be described. A sector consists of a head and, optionally, a body. The head is the first link of the sector which fully exploits both DoFs and the body with the rest of the links, while each one has $\phi_i=0$  and the same $\theta_i$ (Fig \ref{fig_sector}). For example a sector $t$ can be fully described by the vector:
\begin{equation}S_t=
 \begin{bmatrix}
     i_t& u_t & \phi_{h_t}& \theta_{h_t} &\theta_{b_t}  
\end{bmatrix}\end{equation}

\noindent where $i_t$ is the first $i$ link of the sector, $u_t$ the number of the links of the body, $\phi_{h_t}$, $\theta_{h_t}$ correspond to the head and $\theta_{b_t}$ corresponds to the body.
\begin{figure}[th]
\centering
\includegraphics[width=0.30\textwidth]{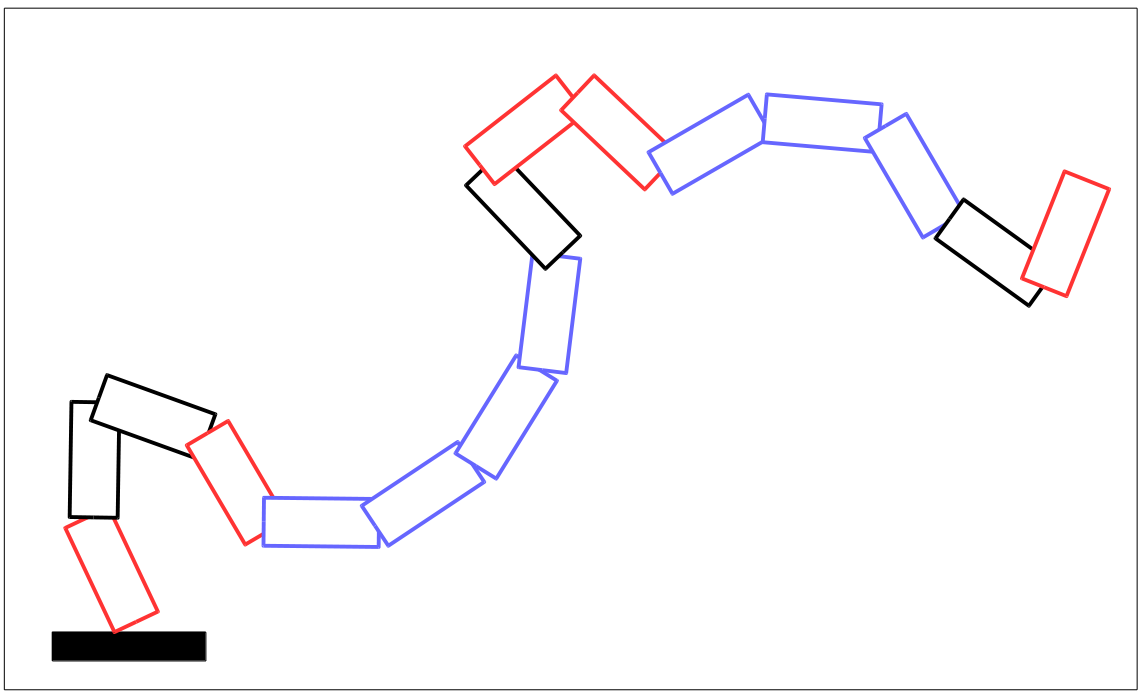}
\caption{A random kinematic state of a hyper\hyp redundant arm of 16 links. In this state (H=\{1,-1,-1,1,0,0,0,0,-1,1,1,0,0,0,-1,1\}) there are 4 damaged links but not only the arm is functional, but also the DoFs are reduced from 32 to 17, leading to shorter response time.}
\label{fig_state}
\end{figure}
Since every link has 2 abilities of movement (head, body) the meta\hyp controlling function is:
\begin{equation}
h:\{1,2,3,...,N\}\rightarrow\{-1,0,1\}
\end{equation}
\noindent where
\begin{equation}
h(i)=
\begin{cases}
 0 & \parbox[t]{.6\columnwidth}{is a link of the body of a sector,}\\
 1 & \parbox[t]{.6\columnwidth}{is the head of a sector,}\\
-1  & \parbox[t]{.6\columnwidth}{is a damaged immobile link with constant  $\Phi_i$, $\Theta_i$.}
\end{cases}
\end{equation}

It is easy now to build the $V$ and the $F$ matrices:
\begin{eqnarray}
&V=
\begin{bmatrix}
     S_1 & S_2 & \dots & S_t & \dots & S_r
\end{bmatrix}^T,\\
&F=
\begin{bmatrix}
     f_1 & f_2 & \dots & f_i & \dots & f_c
\end{bmatrix}, \text{ if } h(f_i)=-1
\end{eqnarray}

From the above formulation notice that the method not only takes care of the damaged links but also reduces the $q$ vector into $Q=[\phi_{h_1}, \theta_{h_1},  \theta_{b_1}, \phi_{h_2}, \theta_{h_2},\dots, \theta_{b_{r-1}}, \phi_{h_r}, \theta_{h_r}, \theta_{b_r}]^T$. Recall that $r$ is the total number of the sectors in the case that every sector has a body, else, the $\theta_{b_t}$ for every sector $t$ without body is missing. So for every body of a sector $t$, regardless the $u_t$ there is only one single variable describing its movement and an entire sector of many links can be described by only 3 joint variables. See Fig. \ref{fig_state} for an example.

To solve the direct and inverse kinematic we produce the homogeneous transformation for each sector separately. First, the homogeneous transformation for the head is:
\begin{equation}A_{h_t}(\phi_{h_t},\theta_{h_t})=A_{1_{i_t}}(\phi_{h_t})A_{2_{i_t}}(\theta_{h_t})\end{equation}
and for the body, since $\phi_i=0$ is:
\begin{equation}
\begin{split}
A_{b_t}(\theta_{b_t})={\displaystyle \prod_{j=1}^{u_t} A_{1_{i_t+j}}(0)A_{2_{i_t+j}}(\theta_{i_t+j})}=\\
{\displaystyle \prod_{j=1}^{u_t} IA_{2_{i_t+j}}(\theta_{i_t+j})}={\displaystyle \prod_{j=1}^{u_t} A_{2_{i_t+j}}(\theta_{i_t+j})}
\end{split}
\end{equation}
It is obvious that the number of matrix multiplications is reduced. Moreover, it can be proved, by implementing the law of cosines iteratively for every link of the body, that if $(u_t-1)\theta_{b_{t}}<2\pi$ the above iteration is not necessary and that product can be replaced with a single matrix:
%
\begin{equation}
\begin{split}
A_{3_t}(\theta_{b_t})=\\
{\footnotesize
 \begin{bmatrix}
     1                  & 0                 &0                  & 0 \\
     0                  & \cos(R'_b)\cos(R_{D_b})-    & \cos(R'_b)\sin(R_{D_b})+    & D{b_t}\sin(R_{D_b}) \\
     					& \sin(R'_b)\sin(R_{D_b})		& \sin(R'_b)\cos(R_{D_b}) & \\
     0                  & -\cos(R'_b)\sin(R_{D_b})-   & \cos(R'_b)\cos(R_{D_b})     & D{b_t}\cos(R_{D_b}) \\
     					&\cos(R'_b)\sin(R_{D_b})	  & -\sin(R'_b)\sin(R_{D_b}) & \\
     0                  & 0                 & 0                 & 1
\end{bmatrix}
}
\end{split}
\label{ll}
\end{equation}

\noindent where:
\begin{eqnarray}
&R_{D_b}=\theta_{b_t}+(u_t-1){\theta_{b_t}\over2}\\
&R'_b=u_t\theta_{b_t}-R_{D_b}\\
&D_{b_t}=
 \begin{cases} 
      d & u_t=1 \\
      x_{u_t} & u_t\geq 2 
   \end{cases}
\end{eqnarray}
\noindent where the $x_{u_t}$ is given by the following recursive function:
\begin{equation}
 x_z=
 \begin{cases} 
      \sqrt[2]{2d^2(1+\cos\theta_{b_t})} & z=2 \\
      \sqrt[2]{x^2_{z-1}+d^2+2x_{z-1}d\cos((u_t+1){\theta_{b_t}\over2})} & z>2 
   \end{cases}
\end{equation}

So, the entire homogeneous transformation for a sector is:
\begin{equation}
\begin{split}
A_{S_t}(\phi_{h_t}, \theta_{h_t}, \theta_{b_t})=A_{h_t}(\phi_{h_t}, \theta_{h_t})A_{b_t}(\theta_{b_t})\\=A_1(\phi_{h_t})A_2(\theta_{h_t})A_3(\theta_{b_t})
\end{split}
\end{equation}if it has body, or:
\begin{equation}
A_{S_t}(\phi_{h_t}, \theta_{h_t})=A_{h_t}(\phi_{h_t}, \theta_{h_t})=A_1(\phi_{h_t})A_2(\theta_{h_t})
\end{equation} without body.

The homogeneous transformation for the $i$ damaged link with constant values $\Phi_i$, $\Theta_i$ is:
\begin{equation}
\begin{split}
B_i(\Phi_i, \Theta_i)=A_{1_i}(\Phi_i)A_{2_i}(\Theta_i)=\\
{\footnotesize
 \begin{bmatrix}
     \cos(\Phi_i)       & \sin(\Phi_i)\cos(\Theta_i)    &\sin(\Phi_i)sin(\Theta_i)      & d\sin(\Phi_i)sin(\Theta_i) \\
     -\sin(\Phi_i)      & \cos(\Phi_i)\cos(\Theta_i)    & \cos(\Phi_i)\sin(\Theta_i)    & d\cos(\Phi_i)\sin(\Theta_i) \\
     0                  & -\sin(\Theta_i)               & \cos(\Theta_i)                & d\cos(\Theta_i) \\
     0                  & 0                             & 0                             &  1
\end{bmatrix}
}
\end{split}
\end{equation}
The transformation of the end effector can be easily produced by the following formula:
\begin{equation}
\begin{split}
T^0_N(Q)={\displaystyle \prod_{l=1}^{i_1-1} B_l(\Theta_l, \Phi_l)}{\displaystyle \prod_{t=1}^{r} \{A_{S_t}(\phi_{h_t}, \theta_{h_t}, \theta_{b_t}^*){\displaystyle \prod_{k=i_t+u_t}^{i_{t+1}-1} B_k(\Theta_k, \Phi_k)}\}}\\{\displaystyle \prod_{l=i_r+u_r}^{l=n} B_l(\Theta_l, \Phi_l)}
\end{split}
\label{Tpppp}
\end{equation}
Note that the $\theta_{b_t}$ variable is not defined for sectors without body. The Jacobian matrix is constructed as:
\begin{equation}J=
\begin{bmatrix}
     J_1 & J_2 & \dots & J_t & \dots & J_r
\end{bmatrix}
\end{equation} where if we define $i_{r+1}=N+1$ then:
\begin{equation} J_t=
\begin{cases}   
    \begin{bmatrix}
     J_{P_{1{i_t}}} & J_{P_{2{i_t}}} &  J{P_{i_{t+1}-1}}\\
     J_{O_{1{i_t}}} & J_{O_{2{i_t}}} &  J{O_{i_{t+1}-1}}
    \end{bmatrix} & \text{if t has a body}\\
    \begin{bmatrix}
     J_{P_{1{i_t}}} & J_{P_{i_{t+1}-1}} \\
     J_{O_{1{i_t}}} & J_{O_{i_{t+1}-1}} 
    \end{bmatrix}  & \text{otherwise}
\end{cases}
\end{equation}
and without considering damaged links:
\begin{equation}
\begin{split}
T^0_N(Q)=\prod_{t=1}^{r} \{A_{S_t}(\phi_{h_t}, \theta_{h_t}, \theta_{b_t}^*)
\end{split}
\label{TTTpppp}
\end{equation}
\begin{equation} J_t=
\begin{cases}   
    \begin{bmatrix}
     J_{P_{1_t}} & J_{P_{2_t}} &  J_{P_{3_t}}\\
     J_{O_{1_t}} & J_{O_{2_t}} &  J_{O_{3_t}}
    \end{bmatrix} & \text{if t has a body}\\
    \begin{bmatrix}
     J_{P_{1_t}} & J_{P_{2_t}} \\
     J_{O_{1_t}} & J_{O_{2_t}} 
    \end{bmatrix}  & \text{otherwise}
\end{cases}
\end{equation}

Given the $V$ vector and the $h$ meta\hyp controlling function it is trivial to make the correspondence between the $Q$ vector of the joint variables of our kinematic structure and the $q$ vector of the joint variables of the real structure.

With Equation (\ref{Tpppp}), this hyper\hyp redundant arm can be controlled fully dynamically with respect to the simple definition of a sector. Now, an example of a simple meta\hyp controller, is presented, that aims to control efficiently the arm assuming that there is no need to take care of damaged links. The main idea is to begin with the minimum number of sectors and in case of failure (when the method is not able to find a solution) split in half every sector until every link is a head of a sector. For this purpose a sequence $k$ is used, which returns the maximum number of link in a body, defined as $k_{a+1}=[\frac{k_a}{2}]$ and $k_0=N$ until $k_a=1$, where $a$ is the number of the failures. Then the definition of the $h$ is:
\begin{equation} h_a(i)=
\begin{cases}   
     1 & i=1 \text{ or } H_{a-1}(i)=1\\
     1 & H_{a-1}(i-1)=H_{a-1}(i+1)=0 \text{ and } i\neq N\\
     1 & (i-1)\text{mod } k_a=0\\
     0 & \text{otherwise}
\end{cases}
\label{metameta}
\end{equation}
By this definition it is ensured that every sector has a body until the last state and that when a link becomes head, it stays in that state. Otherwise, since the assumption that every body has $\phi_i=0$ is violated, $q_{2i}$ should be set to $0$ in the end and align them with respect to the other links of the body, so that states are consistent. Also, it is important to mention that this meta\hyp controller can produce [$\log_{2}N$] different kinematic structures (states) of the robotic arm.

We are going to use the assumptions made in this example, with the optimized calculations in Equation \ref{ll} and the meta\hyp controlling function in Equation \ref{metameta}, to compare experimentally the proposed algorithm to the classic approach in the next section.

\section{Experimental Results}
\label{sec:results}

The experimental setup was developed in MATLAB and tested on an Intel\textsuperscript{\textregistered} Core\textsuperscript{TM}2 Duo Processor (2.13 GHz) with 4 GB RAM. 
The hyper\hyp redundant robotic arm in Section~\ref{sec:impl_ex} was implemented, with the definitions of the sectors and the meta\hyp controlling function. The aim of the experiments was to show the advantage of our algorithm, compared to the classic one, in terms of efficiency and processing time for a task. The computational time needed for solving the direct and inverse kinematics problem for a single move was measured using kinematic states from $1$ to $\log_2 N$ of the proposed method and using the classic method. We ran experiments considering hyper\hyp redundant arms with different number of links, ranged from 16 to 200,000 DoFs. 

\begin{figure}[t]
\centering
\includegraphics[width=0.35\textwidth]{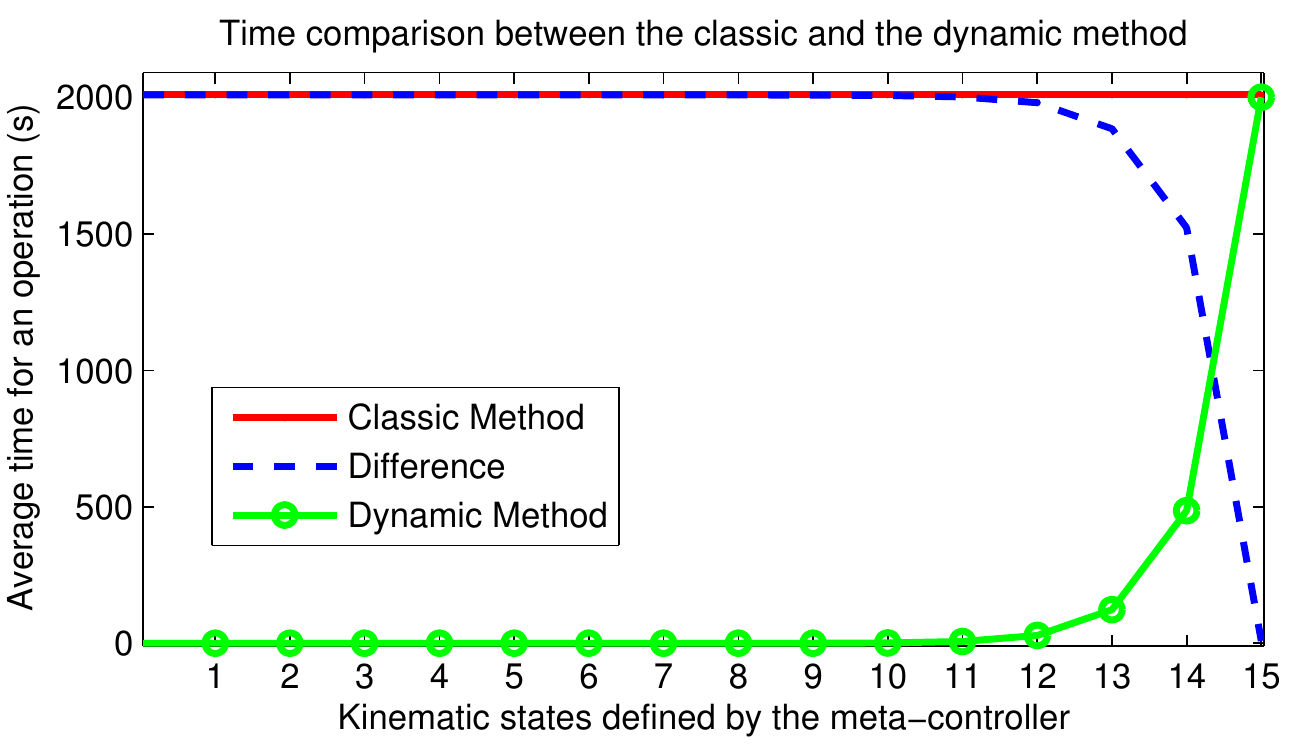}
\caption{Comparison of the processing time needed for a single move between the dynamic and the classic method. Each state represents a different kinematic state of the manipulator according to the meta\hyp controller described in Equation \ref{metameta}.} 
\label{fig_exp}
\end{figure}

Fig. \ref{fig_exp} shows the results for the proposed method and the classic method for a robotic arm of 100,000 links and 200,000 DoFs---similar results were obtained with a different number of links and DoFs. 
The experiments showed clearly that in the worst case, the manipulator was as efficient as in the classic method. Moreover, not only the benefits were growing with more complex and bigger hyper\hyp redundant arms, but also, in the medial case the time needed for an operation was exponentially less than the classic method. For example in the $8^{th}$ state shown in Fig. \ref{fig_exp} our method process time was less than 0.13 sec while for the classic method it was more than 2,000 sec. 

When using the proposed method, it is important to carefully consider the tradeoff between the computational efficiency and the reduction of the DoFs of the hyper\hyp redundant arm. It should be noted that computational needs grow quadratically to the number of the DoFs and that with the use of an optimal meta\hyp controller (a meta\hyp controller that uses the least DoFs needed for an operation) our method always has a performance equal to or significantly better than the classic method. It is worth noticing that in the medial case of the robotic arm with the 100,000 links the manipulator uses approximately 750 DoFs that are almost evenly distributed in its body.

\section{Conclusions}
\label{sec:concl}

This paper introduced a novel way of modeling the kinematics to control hyper\hyp redundant arms in a fully dynamic way. We show that the complexity of both the direct and inverse kinematic problem is unaffected by the number of  links and the degrees of freedom. The proposed method does not require any approximations of a macroscopic perspective as done by previous studies.

The proposed method divides the manipulator into sectors and finds an analytical solution, without discretizing the solution. Due to the meta\hyp controller a general way was introduced, to have the ability to combine different efficient geometric approaches for every sector. Also different methods at a macroscopic perspective can be applied with different approximations and approaches, if the transformation of the sectors are known and these sectors are treated as damaged. Additionally, the kinematic structure can be changed at any time, and the process for solving the direct kinematics and the construction of the Jacobian matrix, can be parallelized for every sector. Moreover, the method is fault tolerant to broken links (immobile links) without the need of implementing extra fault tolerant methods, since the damaged links are considered as sectors that cannot move.

We are currently examine the development of more efficient meta\hyp controller functions that take into account efficient obstacle and knot avoidance policies. Furthermore, we study the relation between the efficiency of the proposed method, with respect to the  structure size. Investigating more technical issues such as how the efficiency of our method is affected by the response delays and the bounded angular velocity of each joint,  on a real hyper\hyp redundant robotic arm is also part of our future plans.



\bibliographystyle{template/IEEEtran}
\bibliography{ref}

\end{document}